# Adaptive probabilistic principal component analysis

**Adam Farooq**
Aston University
United Kingdom

**Yordan P. Raykov**
Aston University
United Kingdom

**Luc Evers**
Radboud University
Netherlands

**Max A. Little**
Aston University
United Kingdom
and
Massachussetts Institute of Technology
United States

## Abstract

Using the linear Gaussian latent variable model as a starting point we relax some of the constraints it imposes by deriving a nonparametric latent feature Gaussian variable model. This model introduces additional discrete latent variables to the original structure. The Bayesian nonparametric nature of this new model allows it to adapt complexity as more data is observed and project each data point onto a varying number of subspaces. The linear relationship between the continuous latent and observed variables make the proposed model straightforward to interpret, resembling a locally adaptive probabilistic PCA (A-PPCA). We propose two alternative Gibbs sampling procedures for inference in the new model and demonstrate its applicability on sensor data for passive health monitoring.

## 1 Introduction

Linear dimensionality reduction methods are a mainstay of high dimensional data analysis, due to their simple geometric interpretation and attractive computational properties. Despite their simplistic assumptions, many such techniques remain widely used in practice arguably due to their intuitive assumptions and robustness to model mis-specification. Many of the more flexible visualization and latent feature techniques can be seen as nonlinear extensions to linear Gaussian methods, namely probabilistic principal components analysis (PCA) and factor analysis (FA) [1]. For example, consider project data $\mathbf{y} \in \mathbb{R}^D$ down to a $K^+ < D$ dimensional linear subspace:

$$\mathbf{y} = \mathbf{W}\mathbf{x} + \boldsymbol{\epsilon}, \boldsymbol{\epsilon} \sim \mathcal{N}(0, \boldsymbol{\Sigma}) \qquad (1)$$

with $\mathbf{x} \in \mathbb{R}^{K^+}$ denoting typically Gaussian latent variables; $\mathbf{W} \in \mathbb{R}^{D \times K^+}$ denoting the projection matrix and $\boldsymbol{\epsilon} \sim \mathcal{N}(0, \boldsymbol{\Sigma})$ being zero-mean Gaussian noise. Then, depending on the shape of $\boldsymbol{\Sigma}$ we can recover probabilistic versions of PCA and FA. However, if we drop the assumption of a linear relationship between $\mathbf{x}$ and $\mathbf{y}$, we can obtain methods such as kernel PCA [2] if data is modelled by $\mathbf{y} = f(\mathbf{W}\mathbf{x}) + \boldsymbol{\epsilon}$ for some pre-defined mapping function $f$; Gaussian process latent variable model (GPLVM)[3] if we place a GP prior on that same $f$; generic topographic mapping (GTM) [4] if we infer $f$ using a radial basis function (RBF) network, or more recently, certain variational autoencoders (VAE)[5] if we model $f$ using a multilayer perceptron. There are also many widely used techniques for data visualization which depart from the above mentioned latent variable assumptions such as t-distributed stochastic neighbour embedding (t-SNE)[6], SNE [7], locally linear neighbourhood embedding [8], Isomap [9] and others, but these methods lack the advantages of probabilistic model-based methods.

Remaining in the family of latent variable models for dimensionality reduction, a complementary view is that more complex input spaces can be sub-divided into regions such that within a region we can efficiently project data down onto a linear subspace – hence we assume that data at least locally lies in a lower dimensional linear subspace. One way to implement such an assumption can be using a mixture model where each component is associated with a transformation matrix and component-specific error. [10] first proposed a probabilistic formulation of the mixture of PCAs together with an expectation-maximization (EM) algorithm for efficient inference. In a follow up, [11] extended the approach to the nonparametric setting where the number of underlying linear subspaces could be inferred automatically from the data, together with the number of underlying principle components in each subspace. [12] proposed an alternative Dirichlet process (DP) mixture of PCAs and derived a computationally-efficient *small variance asymptotics* framework for inference in this new model which scales well to larger problems. DP mixtures of PCAs are ideally suited for clustering the input based on the latent subspace of the data, however, they lead to inefficient representations if one is tasked with identifying how well various data points can be projected down into the lower dimensional linear subspace. This is because each data point can belong only to a single subspace the dimensionality of which is modelled parametrically.

In this work we proposed an alternative discrete-continuous latent feature model aimed at inferring the optimal lower dimensional representation of data in a nonparametric way by modelling independently how likely data points are to lie in the same linear subspace. The probabilistic model is obtained by placing a Beta-Bernoulli process prior on the projection matrix $\mathbf{W}$ in (1). The binary matrix defined by the Beta-Bernoulli process can be used to infer groups of points which are likely to lie in the same lower dimensional subspace; effectively each data point is associated with a selected subset of 1-D linear subspaces. In order to enforce sparse encoding of the data onto these subspaces, we also assume that subpaces (encoded in the columns of the projection matrix $\mathbf{W}$ from (1)) are orthogonal. Given that a group of points share exactly the same set of subspaces (which we call *features*), they are effectively modelled by the same probabilistic PCA model [1], hence if the values of $\mathbf{W}$ are all non-zeros the proposed model collapses to classical probabilistic PCA; conversely, if each point is constrained to belong to a single subspace, we recover a 1-D mixture of PCAs. The proposed model was motivated by the nonparametric extension of FA by [13] in which the columns of $\mathbf{W}$ are not assumed orthogonal. It is also related to the linear Gaussian latent variable model from [14, 15] where each feature, instead of specifying a projection vector directly, specifies the Gaussian components that generate the observation. A related nonparametric PCA was also recently proposed in [16] where alternative parametrizations of $\mathbf{W}$ and model noise are assumed which requires more complex inference using Metropolis-Hastings. In this work we have developed both "vanilla" Gibbs and efficient hybrid Gibbs samplers for inference in this proposed *adaptive probabilistic PCA*(A-PPCA) which lead to straightforward and reasonably scalable inference.

## 2 Adaptive probabilistic PCA

Let us denote the input (D × N) matrix with $\mathbf{Y} = [\mathbf{y}_1 \ldots, \mathbf{y}_N]$; to simplify derivations and without loss of generality we can assume that the mean of $\mathbf{Y}$ is zero. We assume that each point $n$ can be described as $\mathbf{y}_n = \mathbf{W}(\mathbf{x}_n \odot \mathbf{z}_n) + \boldsymbol{\epsilon}_n$ where $\mathbf{x}_n$ is a $\mathrm{K}^+$-dimensional latent variable, $\mathbf{W} = [\mathbf{w}_1, \mathbf{w}_2, ..., \mathbf{w}_{\mathrm{K}^+}]$ is an unobserved (D × $\mathrm{K}^+$) projection matrix with $\mathbf{w}_i \perp \mathbf{w}_j \forall i \neq j$, $\mathbf{z}_n$ is a binary indicator vector denoting the active subspaces for point $n$, $\odot$ denotes the Hadamard product and $\boldsymbol{\epsilon}_n$ is zero-mean Gaussian noise with covariance $\sigma_y^2 \mathbb{I}_D$. Equivalently, we can also write $\mathbf{W}(\mathbf{x}_n \odot \mathbf{z}_n) = \mathbf{W}\mathbf{A}^{(n)}\mathbf{x}_n$, where $\mathbf{A}^{(n)} = \left(\mathbb{I}_{\mathrm{K}^+} \odot \mathbf{z}_n \mathbf{z}_n^T\right)$ which holds for all binary vectors $\mathbf{z}_n$. Under the proposed model we can write the likelihood of point $n$ as:

$$\mathrm{P}(\mathbf{y}_n \mid \mathbf{W}, \mathbf{x}_n, \mathbf{A}^{(n)}, \sigma_y) = \frac{1}{\left(2\pi\sigma_y^2\right)^{D/2}} \exp\left(-\frac{1}{2\sigma_y^2}(\mathbf{y}_n - \mathbf{W}\mathbf{A}^{(n)}\mathbf{x}_n)^{\mathrm{T}}(\mathbf{y}_n - \mathbf{W}\mathbf{A}^{(n)}\mathbf{x}_n)\right) \quad (2)$$

**Distribution of matrix X:** $\mathbf{X} = [\mathbf{x}_1, \ldots, \mathbf{x}_n]^T$ is an (N × $\mathrm{K}^+$) matrix, the rows of which are drawn from independent multivariate Gaussians (with mean zero and covariance matrix $\sigma_x^2 \mathbb{I}_{\mathrm{K}^+}$).

**Distribution of matrix Z:** $\mathbf{Z} = [\mathbf{z}_1, \ldots, \mathbf{z}_n]^T$ is an $(N \times K^+)$ matrix and contains the latent feature indicators denoting under which of the $K^+$ number of 1-D subspaces an observation is associated. We can use an Indian buffet process (IBP) as a prior over the latent feature indicators $\mathbf{Z}$ and the number of features $K^+$, the IBP is a marginal process of the Beta-Bernoulli process and can be described using the following "restaurant" analogy: (1) the first customer enters the buffet and takes the first Poisson $(\alpha)$ number of dishes ($\alpha$ is a concentration parameter) and (2) the $n^{\text{th}}$ customer runs along the buffet and selects a previously sampled dish with probability $\frac{m_k}{n}$, where $m_k$ is the number of people who have already sampled dish $k$; once the $n^{\text{th}}$ customer reaches the end of the buffet they then take Poisson $\left(\frac{\alpha}{n}\right)$ number of new dishes. Identifying customers with observations, and dishes with 1-D subspaces, then the IBP can be used to derive straightforward inference for $\mathbf{Z}$ (see [17] for more detail).

**Distribution of projection matrix W**: Once the IBP samples a new direction $K^+ + 1$, we must sample the corresponding projection vector for the matrix $\mathbf{W} = [\mathbf{w}_1, \mathbf{w}_2, ..., \mathbf{w}_{K^+}]$, while ensuring it is orthonormal to the existing $K^+$ directions, i.e. we have $(D - K^+)$ degrees of freedom to choose the next direction. Let $\mathbf{B} = [\mathbf{b}_1, ..., \mathbf{b}_{D-K^+}]$ be a $(D \times (D - K^+))$ matrix containing the orthonormal basis of all $(D - K^+)$ directions which are orthogonal to the matrix $\mathbf{W}$, we then draw a $(D - K^+)$-dimensional unit vector $\mathbf{v}_{K^++1}$ from the Bingham distribution with parameter $\left(\sigma_v \mathbf{B}^T \mathbf{Y} \mathbf{Y}^T \mathbf{B}\right)$ such that $\text{P}(\mathbf{v}_{K^++1}) \propto \exp\left(\mathbf{v}_{K^++1}^T \left(\sigma_v \mathbf{B}^T \mathbf{Y} \mathbf{Y}^T \mathbf{B}\right) \mathbf{v}_{K^++1}\right)$ (see [18] for more details) and $\sigma_v$ is a scalar hyper-parameter; a Bingham prior was selected because it is conjugate with the likelihood in (3). The update for the $(K^+ + 1)^{\text{th}}$ direction is $\mathbf{w}_{K^++1} = \mathbf{B}\mathbf{v}_{K+1}$ and we write the updated projection matrix as $\mathbf{W} = [\mathbf{w}_1, \mathbf{w}_2, ..., \mathbf{w}_{K^+}, \mathbf{w}_{K^++1}]$. In Appendix A we list a simple Gibbs sampling procedure for inference in this model obtained after marginalizing over $\mathbf{X}$. We also propose a more scalable hybrid sampling procedure which assumes a non-informative prior over $\mathbf{W}$ and proceeds with expectation updates for the latent $\mathbf{X}$ and maximization updates on $\mathbf{W}$. The method keeps the sampling updates of the indicators $\mathbf{z}_n$ which gives the procedure a good chance of escaping locally optimal solutions. The new update of $\mathbf{W}$ can be performed in parallel which can lead to significant computational reductions for large D. The detailed procedure can be found in Appendix C and we also show summarized pseudocode in Algorithm 1.

---

**Algorithm 1** Summarized hybrid Gibbs pseudocode for A-PPCA.

**Input:** $\mathbf{Y}, \Theta$, MaxIter;   **Output:** $\mathbf{Z}, \Theta$
**Initialise:** Set $K^+ = 1, \mathbf{w}_1$
**for** iter $\leftarrow 1$ to maxiter
    **for** $n \leftarrow 1$ to N
        **for** $k \leftarrow 1$ to $K^+$
            Sample $z_{kn} \propto \text{P}\left(z_{kn} \mid \mathbf{y}_n, \mathbf{W}, \mathbf{A}^{(n)}, \sigma_y\right)$
        Propose adding Poisson $\left(\frac{\alpha}{N}\right)$ number of new subspaces
        Accept or reject the proposal using a Metropolis-Hastings step
    **for** $n \leftarrow 1$ to N
        Set $\boldsymbol{x}_n = \left(\sigma_x^{-2}\mathbb{I}_{K^+} + \sigma_y^{-2}\mathbf{A}^{(n)}\mathbf{W}^{\text{T}}\mathbf{W}\mathbf{A}^{(n)}\right)^{-1}\left(\sigma_y^{-2}\mathbf{A}^{(n)}\mathbf{W}^{\text{T}}\mathbf{y}_n\right)$
        Set $\boldsymbol{\Psi}_n = \left(\sigma_x^{-2}\mathbb{I}_{K^+} + \sigma_y^{-2}\mathbf{A}^{(n)}\mathbf{W}^{\text{T}}\mathbf{W}\mathbf{A}^{(n)}\right)^{-1} + \boldsymbol{x}_n\boldsymbol{x}_n^T$
    Update $\mathbf{W} = \left(\sum_{n=1}^{N} \mathbf{y}_n \left(\mathbf{A}^{(n)}\boldsymbol{x}_n\right)^{\text{T}}\right)\left(\sum_{n=1}^{N} \mathbf{A}^{(n)}\boldsymbol{x}_n\boldsymbol{x}_n^T\mathbf{A}^{(n)}\right)^{-1}$
    Update $\sigma_y^2 = \frac{1}{ND}\sum_{n=1}^{N}\left(\mathbf{y}_n^{\text{T}}\mathbf{y}_n - 2\boldsymbol{x}_n^T\mathbf{A}^{(n)}\mathbf{W}^{\text{T}}\mathbf{y}_n + \text{Tr}\left(\mathbf{A}^{(n)}\mathbf{W}^{\text{T}}\mathbf{W}\mathbf{A}^{(n)}\boldsymbol{\Psi}_n\right)\right)$
    Update $\sigma_x^2 = \frac{1}{NK^+}\sum_{n=1}^{N}\text{Tr}(\boldsymbol{\Psi}_n)$

---

## 3 Inferring the latent dimensionality of multi-source data in free living

In Appendix B we compare the performance of both the proposed collapsed Gibbs and the hybrid Gibbs samplers with EM for the A-PPCA and the mixture of PCAs models on synthetically generated data. Next, we use the A-PPCA model to identify the relevance of different sensors placed on different parts of the human body when remotely recording different activities. The problem is motivated by the PD@Home validation study in which a small group of individuals (25 Parkin-son's disease (PD) patients and 25 controls) are monitored continuously for periods of time using wearable devices attached to different parts of their body. Using A-PPCA, the aim is to reduce the dimensionality of this high-resolution sensor data to simplify further storage and processing. Data is taken from a single healthy subject using accelerometer sensors worn on both ankles, both wrists and the lower back. Each accelerometer sensor produces 3-D orthogonal $x, y, z$ axis outputs at 120Hz sampling frequency for the duration of 45 minutes. This sensor data is concatenated into a single $15 \times 324,000$ matrix. Depending on the hyperparameters of the model one can control the resolution and the extent to which latent features are identified from the observed sensor data. In Figure 1 we optimized the model likelihood to display the estimated feature (transformation) matrix $\mathbf{W}$ where $K^+ = 3$ latent features were identified. Looking at the absolute values of the features, we learn that data points which share feature 1 are mostly affected by the variance of dimensions 4 & 6 ($x$ and $z$ axis of right ankle); points which share feature 2 are mostly impacted by dimensions 1 & 3 ($x$ and $z$ axis of left ankle) and points which share feature 3 are mostly impacted by dimensions 2,4,5,10 & 14 ($y$ axis for left and right ankle; $x$ axis for the right wrist; $y$ axis for the lower back). Changing the hyperparameters to identify more features leads to features dominated by other body locations. Using the estimated $\mathbf{W}$ and $\mathbf{Z}$ we can adaptively choose the most non-overlapping, active sensors, identify limbs used for the particular activities and store or transmit a significantly more efficient representation of the data. In Table 1 we also plot the reconstruction error obtained when using the estimated $\mathbf{W}$ to reduce and recover $\mathbf{Y}$. The reconstruction of $\mathbf{Y}$ can be seen in Figure 2 in the Appendix.

Table 1: Performance measures obtained using by fitting A-PPCA model using hybrid Gibbs on the PD@Home dataset for different $\alpha$. See Appendix C for clarification of orthonormality loss for across the columns of $\mathbf{W}$.

| IBP parameter $\alpha$ | 0.1 | 0.2 | 0.3 | 0.4 |
|---|---|---|---|---|
| Number of inferred features $K^+$ | 3 | 5 | 6 | 8 |
| Mean absolute error | 0.088 | 0.075 | 0.067 | 0.059 |
| Orthonormality loss | $7.35°$ | $18.91°$ | $7.42°$ | $8.26°$ |

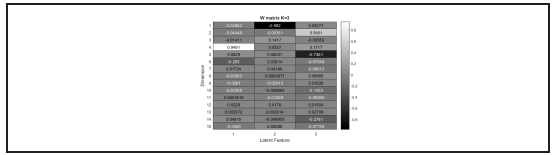

Figure 1: Heatmap of the rotation matrix $\mathbf{W} = [\mathbf{w}_1, \mathbf{w}_2, \mathbf{w}_3]$ (when $K^+ = 3$).

## 4 Conclusion

We have presented an adaptive nonparametric extension of the ubiquitous probabilistic PCA algorithm (A-PPCA). Both collapsed and hybrid Gibbs sampling schemes are proposed for inference in this A-PPCA model and we compare them on synthetically generated data. This new model extends discrete-continuous dimensionality reduction latent variable models to large-scale data problems. The proposed model and inference algorithms can be extended in many ways. For example, it would be straightforward to use a fully deterministic sampling inference (for example, memoized variational inference [19]) and also explore combinations of discrete latent feature models with nonlinear latent variable models such as GPLVMs. For example, introducing nonparametric latent features to GPLVMs is likely to significantly reduce the effort required in choosing a complex kernel function and enable flexible models with a combination of highly interpretable components.

# References

[1] M. E. Tipping and C. M. Bishop. Probabilistic principal component analysis. *Journal of the Royal Statistical Society: Series B (Statistical Methodology)*, 61(3):611–622, 1999. 1, 1, A

[2] B. Schölkopf, A. Smola, and K.-R. Müller. Kernel principal component analysis. In *International Conference on Artificial Neural Networks*, pages 583–588. Springer, 1997. 1

[3] N. D. Lawrence. Gaussian process latent variable models for visualisation of high dimensional data. In *Advances in neural information processing systems*, pages 329–336, 2004. 1

[4] C. M. Bishop, M. Svensén, and C. K. Williams. Gtm: The generative topographic mapping. *Neural computation*, 10(1):215–234, 1998. 1

[5] D. P. Kingma and M. Welling. Auto-encoding variational bayes. *arXiv preprint arXiv:1312.6114*, 2013. 1

[6] L. v. d. Maaten and G. Hinton. Visualizing data using t-sne. *Journal of machine learning research*, 9(Nov):2579–2605, 2008. 1

[7] G. E. Hinton and S. T. Roweis. Stochastic neighbor embedding. In *Advances in neural information processing systems*, pages 857–864, 2003. 1

[8] D. L. Donoho and C. Grimes. Hessian eigenmaps: Locally linear embedding techniques for high-dimensional data. *Proceedings of the National Academy of Sciences*, 100(10):5591–5596, 2003. 1

[9] J. B. Tenenbaum, V. De Silva, and J. C. Langford. A global geometric framework for nonlinear dimensionality reduction. *science*, 290(5500):2319–2323, 2000. 1

[10] M. E. Tipping and C. M. Bishop. Mixtures of probabilistic principal component analyzers. *Neural computation*, 11(2):443–482, 1999. 1

[11] M. Chen, J. Silva, J. Paisley, C. Wang, D. Dunson, and L. Carin. Compressive sensing on manifolds using a nonparametric mixture of factor analyzers: Algorithm and performance bounds. *IEEE Transactions on Signal Processing*, 58(12):6140–6155, 2010. 1

[12] Y. Wang and J. Zhu. Dp-space: Bayesian nonparametric subspace clustering with small-variance asymptotics. In *International Conference on Machine Learning*, pages 862–870, 2015. 1

[13] J. Paisley and L. Carin. Nonparametric factor analysis with beta process priors. In *Proceedings of the 26th Annual International Conference on Machine Learning*, pages 777–784. ACM, 2009. 1

[14] Z. Ghahramani and T. L. Griffiths. Infinite latent feature models and the indian buffet process. In *Advances in neural information processing systems*, pages 475–482, 2006. 1

[15] Z. Ghahramani, T. L. Griffiths, and P. Sollich. Bayesian nonparametric latent feature models. 2007. 1

[16] C. Elvira, P. Chainais, and N. Dobigeon. Bayesian nonparametric principal component analysis. *arXiv preprint arXiv:1709.05667*, 2017. 1

[17] T. L. Griffiths and Z. Ghahramani. The indian buffet process: An introduction and review. *Journal of Machine Learning Research*, 12(Apr):1185–1224, 2011. 2

[18] C. Bingham. An antipodally symmetric distribution on the sphere. *The Annals of Statistics*, pages 1201–1225, 1974. 2

[19] M. C. Hughes and E. Sudderth. Memoized online variational inference for dirichlet process mixture models. In *Advances in Neural Information Processing Systems*, pages 1133–1141, 2013. 4

## A Collapsed Gibbs Pseudocode

**Marginalise over latent values:** We can simplify the model by marginalising over the latent values $\mathbf{x}_n$, such that

$$P(\mathbf{y}_n|\mathbf{W}, \mathbf{A}^{(n)}, \sigma_X, \sigma_y) = \int P(\mathbf{y}_n \mid \mathbf{W}, \mathbf{x}_n, \mathbf{A}^{(n)}, \sigma_y) \times P(\mathbf{x}_n|\sigma_x) d\mathbf{x}_n \quad (3)$$

$$= \frac{|\mathbf{C}_n|^{-1/2}}{(2\pi)^{D/2}} \exp\left\{-\frac{1}{2}\mathbf{y}_n^T \mathbf{C}_n^{-1} \mathbf{y}_n\right\}$$

where $\mathbf{C}_n = \left((\mathbf{W}\mathbf{A}^{(n)})(\mathbf{W}\mathbf{A}^{(n)})^T \sigma_x^2 + \sigma_y^2 \mathbb{I}_D\right)$, one can observe the similarities between (3) and eq (6) from [1]; this similarity occurs because the probabilistic Principle component analysis (PPCA) model is a special case of the A-PPCA model when the variance over the latent features is one ($\sigma_x^2 = 1$) and the matrix $\mathbf{Z} = [\mathbf{z}_1, \ldots, \mathbf{z}_n]^T$ is full of ones implying all observations are active in all $K^+$ number of 1-dimensional subspaces; and PCA is a special case of the PPCA model (see section 3.3 of [1]).

We should also observe the following updates on the matrix $\mathbf{Z}$

$$P(z_{kn} = 1 \mid \mathbf{y}_n, \mathbf{W}, \mathbf{A}^{(n)}, \sigma_y) \propto \frac{m_{k,\neg n}}{N} \times P(\mathbf{y}_n \mid \mathbf{W}, A_{k,k}^{(n)} = 1, \mathbf{A}_{\neg k, \neg k}^{(n)}, \sigma_y)$$

$$P(z_{kn} = 0 \mid \mathbf{y}_n, \mathbf{W}, \mathbf{A}^{(n)}, \sigma_y) \propto \frac{N - m_{k,\neg n}}{N} \times P(\mathbf{y}_n \mid \mathbf{W}, A_{k,k}^{(n)} = 0, \mathbf{A}_{\neg k, \neg k}^{(n)}, \sigma_y)$$

where $m_{k,\neg n} = \sum_{i \neq n} \mathbf{z}_{ki}$.

**Collapsed Gibbs pseudocode:** The collapsed Gibbs pseudocode can be seen below in Algorithm 3 where Bingham $(\cdot)$ is the Bingham distribution, and Orth $(\cdot)$ represents the orthornomal basis of a matrix.

**Hybrid Gibbs pseudocode:** We can increase the computational performance of the collapsed Gibbs sampler in Algorithm 2 by lifting the orthonormal constraint of the columns of the matrix $\mathbf{W}$ and maximising it with respect to the model likelihood, the pseudocode of this method is highlighted in Algorithm 3.

## B Synthetic study

We generated synthetic data which followed the following set up $\mathbf{Y} = \mathbf{W}(\mathbf{X} \odot \mathbf{Z}) + \boldsymbol{\epsilon}$, where $\mathbf{Y}$ is a ($D \times 1000$) observational matrix, $\mathbf{X}$ is a ($K^+ \times 1000$) matrix which has each column drawn from a multivariate Gaussian with mean zero and covariance matrix $(\sigma_x^2 \mathbb{I}_{K^+})$, where $\sigma_x$ is some parameter set to 1.5, $\mathbf{Z}$ is a random ($K^+ \times 1000$) binary matrix which has at least one 1 in each column, the matrix $\mathbf{W}$ is ($D \times K^+$) projection matrix with perpendicular columns, and $\boldsymbol{\epsilon}$ is a ($15 \times 1000$) noise matrix which has each column drawn from a multivariate Gaussian with mean zero and covariance matrix $(\sigma_y^2 \mathbb{I}_D)$, where $\sigma_y$ is scalar parameter set to 0.5. We compare the mean absolute error of the PCA, MPPCA and the A-PPCA with varying values of D and $K^+$.

Table 2: Mean absolute error for PCA, MPPCA and A-PPCA with $D = 15$ over different values of $K^+$

| Latent subspace dimension $K^+$ | 1 | 2 | 4 | 7 | 9 |
|---|---|---|---|---|---|
| PCA | 0.255 | 0.251 | 0.249 | 0.257 | 0.249 |
| MPPCA | 0.255 | 0.276 | 0.308 | 0.387 | 0.425 |
| Collapsed Gibbs for A-PPCA | 0.237 | 0.236 | 0.221 | 0.213 | 0.184 |
| Hybrid Gibbs for A-PPCA | 0.186 | 0.172 | 0.149 | 0.101 | 0.090 |

**Algorithm 2** Collapsed Gibbs pseudocode for A-PPCA
---
**Input:** $\mathbf{Y}, \Theta, \text{MaxIter}$
**Initialise:** Set $K^+ = 1$
**for** iter $\leftarrow 1$ to MaxIter
  **if** iter=1
    Sample $\mathbf{v}_1 \sim \text{Bingham}\left(\sigma_v \mathbf{Y}\mathbf{Y}^T\right)$
    Set $\mathbf{w}_1 = \mathbf{v}_1$
  **end**
  **for** $n \leftarrow 1$ to N
    **for** $k \leftarrow 1$ to $K^+$
      Sample $z_{kn} \propto P\left(z_{kn} \mid \mathbf{y}_n, \mathbf{W}, \mathbf{A}^{(n)}, \sigma_y\right)$
    **end**
    Sample $\kappa = \text{Poisson}\left(\frac{\alpha}{n}\right)$
    **if** $\kappa > 1$
      Set $\mathbf{W}_{\text{prop}} = \mathbf{W}$
      $K^+_{\text{prop}} = K^+$
      **for** $i \leftarrow 1$ to $\kappa$
        Set $K^+_{\text{prop}} = K^+_{\text{prop}} + 1$
        Set $\mathbf{B} = \text{Orth}\left(\mathbf{W}^\perp_{\text{prop}}\right)$
        Sample $\mathbf{v}_{K^+} \sim \text{Bingham}\left(\sigma_v \mathbf{B}^T \mathbf{Y}\mathbf{Y}^T \mathbf{B}\right)$
        Set $\mathbf{w}_{K^+} = \mathbf{B}\mathbf{v}_{K^+}$
        Update $\mathbf{W}_{\text{prop}} = [\mathbf{W}_{\text{prop}}, \mathbf{w}_{K^+}]$
      **end**
      Accept or reject $\left(\mathbf{W}_{\text{prop}}, K^+_{\text{prop}}\right)$ as $\left(\mathbf{W}, K^+\right)$ using a Metropolis-Hastings step
    **end**
  **end**
  Sample $\mathbf{v}_1 \sim \text{Bingham}\left(\sigma_v \left(\sum_{n=1}^N \mathbf{y}_n \mathbf{y}_n^T\right) + \frac{\sigma_x^2}{2\sigma_y^2(\sigma_x^2 + \sigma_y^2)} \left(\sum_{n=1}^N z_{n1} \mathbf{y}_n \mathbf{y}_n^T\right)\right)$
  Set $\mathbf{w}_1 = \mathbf{v}_1$
  Update $\mathbf{W} = [\mathbf{w}_1]$
  **for** $j \leftarrow 2$ to $K^+$
    Set $\mathbf{B} = \text{Orth}\left([\mathbf{w}_1, \ldots, \mathbf{w}_{j-1}]^\perp\right)$
    Sample $\mathbf{v}_j \sim \text{Bingham}\left(\mathbf{B}^T \left(\sigma_v \left(\sum_{n=1}^N \mathbf{y}_n \mathbf{y}_n^T\right) + \frac{\sigma_x^2}{2\sigma_y^2(\sigma_x^2 + \sigma_y^2)} \left(\sum_{n=1}^N z_{nj} \mathbf{y}_n \mathbf{y}_n^T\right)\right) \mathbf{B}\right)$
    Set $\mathbf{w}_j = \mathbf{B}\mathbf{v}_j$
    Update $\mathbf{W} = [\mathbf{w}_1, \ldots, \mathbf{w}_{j-1}, \mathbf{w}_j]$
  **end**
  Re-sample $\Theta$ using Metropolis-Hastings
**end**
**Output:** $Z, \Theta$
---

Table 3: Mean absolute error for PCA, MPPCA and A-PPCA with $D = 30$ and different values of $K^+$

| Latent subspace dimension $K^+$ | 2 | 6 | 8 | 16 | 18 |
|---|---|---|---|---|---|
| PCA | 0.251 | 0.250 | 0.254 | 0.247 | 0.245 |
| MPPCA | 0.257 | 0.303 | 0.331 | 0.429 | 0.441 |
| Hybrid Gibbs for A-PPCA | 0.156 | 0.134 | 0.121 | 0.099 | 0.088 |

## C  Optimal representation loss

**Orthonormality loss**

We assume the latent subspaces are perpendicular by introducing a constraint on the $(D \times K^+)$ projection matrix $\mathbf{W} = [\mathbf{w}_1, \mathbf{w}_2, ..., \mathbf{w}_{K^+}]$ where $\mathbf{w}_i \perp \mathbf{w}_j \forall i \neq j$ (an angle of $90°$ between all columns of $\mathbf{W}$). However, we may want to partially relax this orthogonality constraint on the matrix $\mathbf{W}$ for computational efficiency. For that purpose, we evaluate how well unconstrained posterior sample of $\mathbf{W}$ meets the conditions of orthogonality. This is done by finding the two columns of the unconstrained matrix $\mathbf{W}$ which have an angle between them furthest from $90°$ (or the two columns which are the least perpendicular); this can be seen in Figure 3.

**Algorithm 3** Hybrid Gibbs pseudocode for A-PPCA

**Input:** $\mathbf{Y}, \Theta, \text{MaxIter}$
**Initialise:** Set $K^+ = 1$
**for** iter $\leftarrow$ 1 to MaxIter
  **if** iter=1
    Sample $\mathbf{w}_1 \sim \text{Uniform}(D)$
  **end**
  **for** $n \leftarrow 1$ to N
    **for** $k \leftarrow 1$ to $K^+$
      Sample $z_{kn} \propto \text{P}\left(z_{kn} \mid \mathbf{y}_n, \mathbf{W}, \mathbf{A}^{(n)}, \sigma_y\right)$
    **end**
    Sample $\kappa = \text{Poisson}\left(\frac{\alpha}{n}\right)$
    **if** $\kappa > 1$
      Set $\mathbf{W}_{\text{prop}} = \mathbf{W}$
      $K^+_{\text{prop}} = K^+$
      **for** $i \leftarrow 1$ to $\kappa$
        Set $K^+_{\text{prop}} = K^+_{\text{prop}} + 1$
        Sample $\mathbf{v}_{K^+} \sim \text{Uniform}(D)$
        Set $\mathbf{w}_{K^+} = \mathbf{v}_{K^+}$
        Update $\mathbf{W}_{\text{prop}} = [\mathbf{W}_{\text{prop}}, \mathbf{w}_{K^+}]$
      **end**
      Accept or reject $\left(\mathbf{W}_{\text{prop}}, K^+_{\text{prop}}\right)$ as $\left(\mathbf{W}, K^+\right)$ using a Metropolis-Hastings step
    **end**
  **end**
  **for** $n \leftarrow 1$ to N
    Set $\boldsymbol{x}_n = \left(\sigma_x^{-2}\mathbb{I}_{K^+} + \sigma_y^{-2}\mathbf{A}^{(n)}\mathbf{W}^{\text{T}}\mathbf{W}\mathbf{A}^{(n)}\right)^{-1}\left(\sigma_y^{-2}\mathbf{A}^{(n)}\mathbf{W}^{\text{T}}\mathbf{y}_n\right)$
    Set $\boldsymbol{\Psi}_n = \left(\sigma_x^{-2}\mathbb{I}_{K^+} + \sigma_y^{-2}\mathbf{A}^{(n)}\mathbf{W}^{\text{T}}\mathbf{W}\mathbf{A}^{(n)}\right)^{-1} + \boldsymbol{x}_n\boldsymbol{x}_n^T$
  **end**
  Update $\mathbf{W} = \left(\sum_{n=1}^N \mathbf{y}_n\left(\mathbf{A}^{(n)}\boldsymbol{x}_n\right)^{\text{T}}\right)\left(\sum_{n=1}^N \mathbf{A}^{(n)}\boldsymbol{x}_n\boldsymbol{x}_n^T\mathbf{A}^{(n)}\right)^{-1}$
  Update $\sigma_y^2 = \frac{1}{ND}\sum_{n=1}^N \left(\mathbf{y}_n^{\text{T}}\mathbf{y}_n - 2\boldsymbol{x}_n^T\mathbf{A}^{(n)}\mathbf{W}^{\text{T}}\mathbf{y}_n + \text{trace}\left(\mathbf{A}^{(n)}\mathbf{W}^{\text{T}}\mathbf{W}\mathbf{A}^{(n)}\boldsymbol{\Psi}_n\right)\right)$
  Update $\sigma_x^2 = \frac{1}{NK^+}\sum_{n=1}^N \left(\text{trace}\left(\boldsymbol{\Psi}_n\right)\right)$
**end**
**Output:** $\mathbf{Z}, \Theta$

Table 4: Mean absolute error for PCA, MPPCA and A-PPCA with $D = 45$ and different values of $K^+$

| Latent subspace dimension $K^+$ | 3 | 9 | 15 | 18 | 27 |
| --- | --- | --- | --- | --- | --- |
| PCA | 0.250 | 0.251 | 0.245 | 0.253 | 0.253 |
| MPPCA | 0.269 | 0.312 | 0.363 | 0.394 | 0.464 |
| Hybrid Gibbs for A-PPCA | 0.141 | 0.129 | 0.114 | 0.106 | 0.097 |

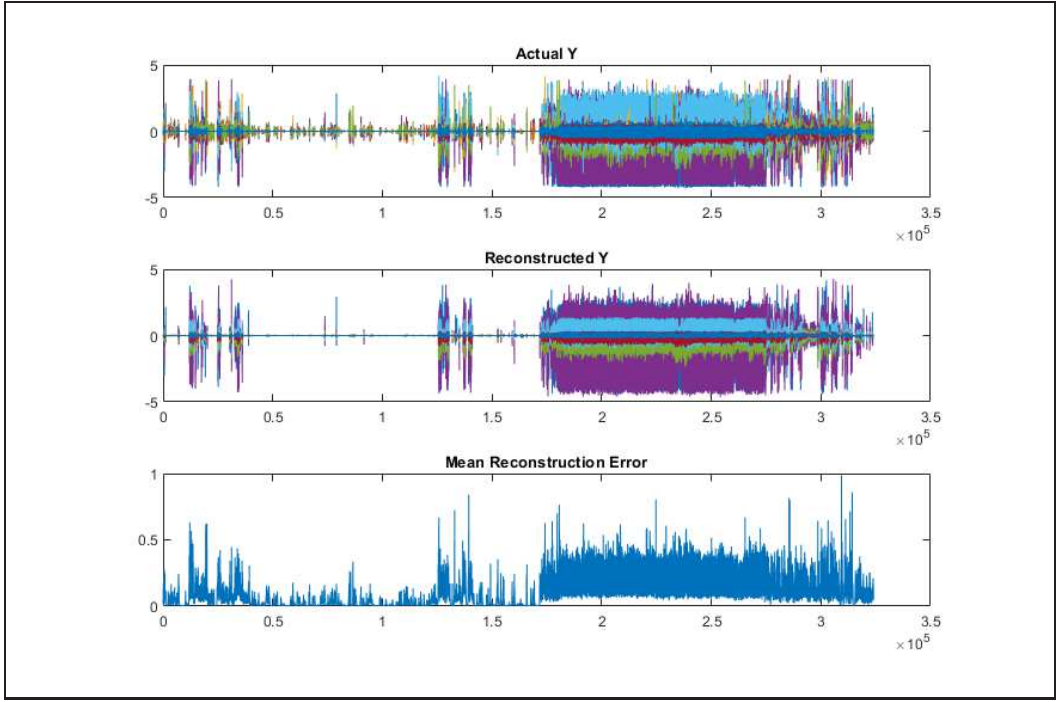

Figure 2: **Top:** Plot of the actual data the $y$ axis represent the values of the sensor data? and the $x$ axis represents time. **Middle:** Plot of the reconstructed data from the top figure. **Bottom:** Plot of the absolute error over all observations

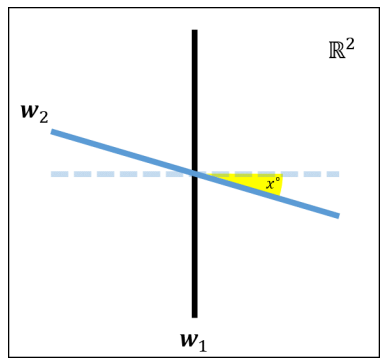

Figure 3: The two columns of the $\mathbf{W} = [\mathbf{w}_1, \mathbf{w}_2]$ matrix represent two directions in $\mathbb{R}^2$, assuming $\mathbf{w}_1$ is fixed and if we impose a perpendicular constraint on the columns of matrix $\mathbf{W}$, then the direction of $\mathbf{w}_2$ would be the dotted blue line, and if we don't impose a perpendicular constraint on the columns of matrix $\mathbf{W}$, then the direction of $\mathbf{w}_2$ would be the solid blue line. To evaluate how well the unconstrained matrix $\mathbf{W}$ meets this perpendicular condition, we find the yellow angle $x°$.